# The Two Sides of the Coin: Hallucination Generation and Detection with LLMs as Evaluators for LLMs

Notebook for the ELOQUENT Lab at CLEF 2024

Anh Thu Maria Bui[1,†], Saskia Felizitas Brech[1,†], Natalie Hußfeldt[1,†], Tobias Jennert[1,†], Melanie Ullrich[1,†], Timo Breuer[1,†], Narjes Nikzad Khasmakhi[1,*,†] and Philipp Schaer[1,†]

[1]*TH Köln – University of Applied Sciences, Cologne, Germany*

**Abstract**

Hallucination detection in Large Language Models (LLMs) is crucial for ensuring their reliability. This work presents our participation in the CLEF ELOQUENT HalluciGen shared task, where the goal is to develop evaluators for both generating and detecting hallucinated content. We explored the capabilities of four LLMs: Llama 3, Gemma, GPT-3.5 Turbo, and GPT-4, for this purpose. We also employed ensemble majority voting to incorporate all four models for the detection task. The results provide valuable insights into the strengths and weaknesses of these LLMs in handling hallucination generation and detection tasks.

## 1. Introduction

The prevalence of large language models advancements and groundbreaking results for many NLP research problems [1], tremendously changed how we generally approach everyday tasks but also how we approach larger more complex problems. The LLMs' ability to combine vast amounts of knowledge from different sources is unparalleled and for some tasks exceeds what humans are able to achieve.

However, blind faith in the generated outputs of these models is critical as they may produce incorrect facts, also known as hallucinations. These false facts can be misleading and are one of the main barriers to using LLMs reliably and in a trustworthy way.

To this end, this work is part of our participation in the ELOQUENT Lab 2024 at CLEF. More specifically, we participate in the HalluciGen task that evaluates if the LLMs themselves are able to correctly detect hallucinations in both human- and machine-generated contexts [2]. The HalluciGen task is divided into two phases over two years. The first phase is dedicated to the builder task, while the second phase will focus on the breaker task. This study specifically targets the first phase where the goal is to create multilingual and monolingual hallucination-aware models. These models are designed to generate and detect 'hallucinated content' in two scenarios: machine translation and paraphrase generation. Figure 1 illustrates an overview of hallucination generation and detection tasks as described by the lab.

The remainder of this work is structured as follows. Section 2 describes our methodology in more detail. Section 3 details the implementation. Section 4 describes our results. Finally, Section 5 concludes our contributions.

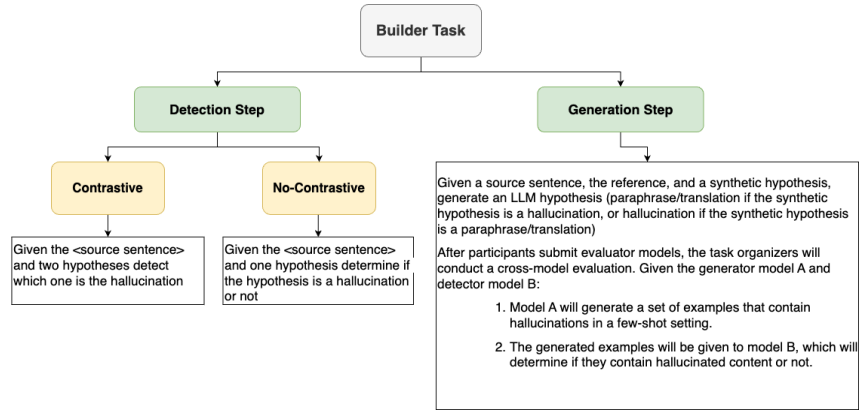

**Figure 1:** An overview of hallucination generation and detection tasks.

## 2. Methodology

This section is divided into two parts: generation and hallucination detection tasks. Before delving into the details of our methodology, it is important to note that prior to receiving the dataset from the organizers, we began familiarizing ourselves with the overall task by applying the three models *Falcon* [3], *MPT* [4], and *Llama 2* [5] to the hallucination detection task on the *SHROOM* dataset [6]. Since the results from these three techniques were unsatisfactory, we excluded them from our implementation for Eloquent Lab.

The models we applied to the Eloquent's dataset in generation and detection tasks were:

- *Meta-Llama/Meta-Llama-3-8B-Instruct* [7, 8]
- *GPT-3.5 Turbo* [9]
- *GPT-4* [10]
- *Google/GEMMA-7B-IT* [11, 12]

We leveraged a combination of open-source and closed-source models. This allows us to evaluate the quality of outputs across different models. Additionally, utilizing open-source models helped us optimize costs. Therefore, we initially experimented with various prompts for the tasks using open-source LLMs to identify the most effective ones. Then, we applied these optimized prompts to closed-source GPT models. Additionally, we did our best to enhance our prompting effectiveness using the *guidance* framework [13].

### 2.1. Hallucination generation task

The task of hallucination generation is divided into two scenarios: machine translation and paraphrasing. The goal of the generation step is to take a source sentence and generate two LLM hypotheses: one that is a correct translation/paraphrase of the source and one that is a hallucinated translation/paraphrase of the source.

Figure 2 indicates the overview of our approach for the generation task. To conduct this task, we took advantage of 'GPT-3.5 Turbo', 'GEMMA-7B-IT', and 'Llama 3'.

### 2.2. Hallucination detection task

The hallucination detection task is to present the LLM with a source sentence and two hypotheses (hyp1 and hyp2) and to determine which hypothesis is a hallucination and which is factually accurate. Our approach involved using four different LLMs, 'GPT-3.5 Turbo', 'Google/GEMMA-7B-IT', 'Llama 3', and 'GPT-4', as classifiers. Additionally, we employed a voting approach as a simple technique of ensemble learning [14] to combine the outputs of these four models.

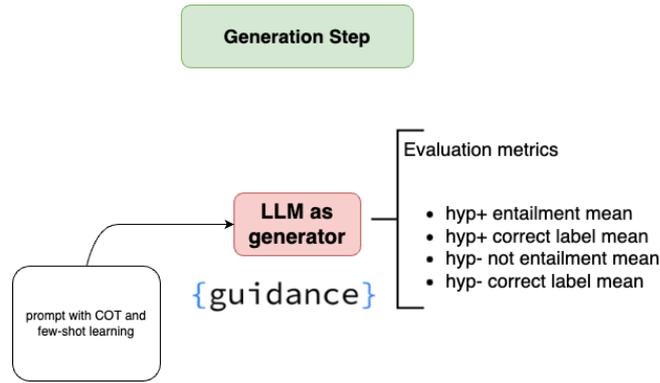

**Figure 2:** An overview of our approach for the generation task.

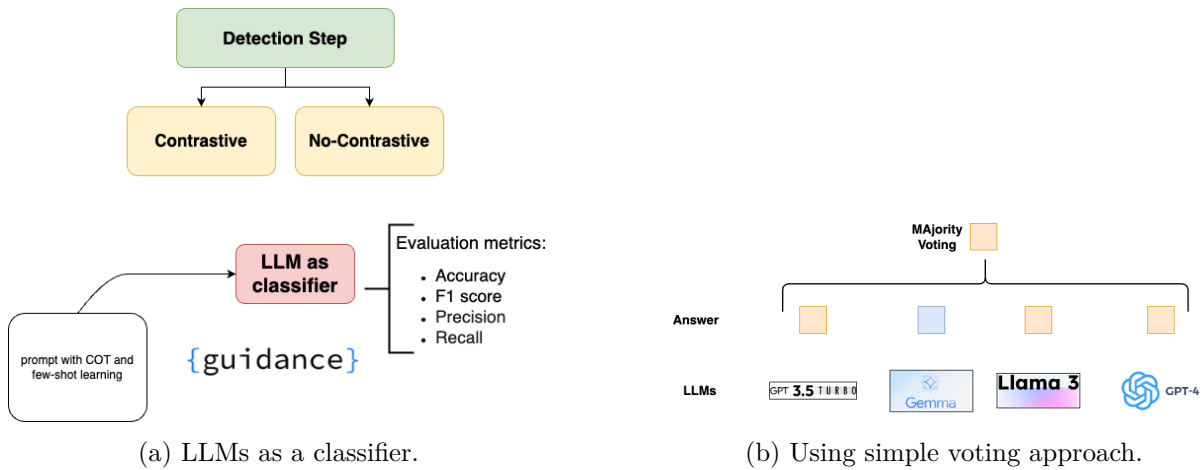

(a) LLMs as a classifier.  (b) Using simple voting approach.

**Figure 3:** An overview of our approach for the detection task.

Furthermore, we experimented with four distinct prompting techniques to provide better guidance to the LLMs and enhance their ability to discriminate between factual and hallucinated information.

- *Type1: Simple Prompt*: Using labels 'hallucinated' or 'not hallucinated'.
- *Type2: Complex Prompt with* 0/1 *Labels*: Specifying the task with labels 0 or 1.
- *Type3: Prompt with Definition and Examples*: Including a definition of hallucination alongside examples labeled 0 and 1.
- *Type4: Prompt with Full Task Description*: Describing the entire translation task (for instance) and hallucination detection goal. Combined Prompt: Combining all the above elements.

## 3. Implementation

This part primarily focuses on how we prompted LLMs, along with the challenges and observations we encountered during the task. We divide this section into three parts: generation, detection, and cross-evaluation tasks.

### 3.1. Generation task

The generation task includes test sets for both paraphrasing and translation tasks.

### 3.1.1. Paraphrasing Generation Task

The paraphrasing generation task involved datasets in English and Swedish, comprising 118 samples for English and 76 samples for Swedish.

The performance of different models, including 'Gemma', 'GPT-3.5 Turbo', and 'Llama 3', was evaluated based on their ability to generate paraphrases for English and Swedish datasets. In the appendix in Section A, Figures 4 to 7 show comprehensive lists of all prompts used for the different models. The following demonstrates some of our observations regarding the implementation of the generation task:

- The performance of the 'Gemma' model varied significantly based on the complexity of the prompts used. Simpler prompts yielded better results that highlight the importance of prompt design. Despite this, the model struggled with understanding specific instructions, such as 'generate hallucination'. Additionally, the generation speed was notably slow.
- For 'GPT-3.5 Turbo', one prompt for English and one prompt for Swedish were employed. The generation speed of 'GPT-3.5 Turbo' was significantly faster compared to other models.
- For 'Llama 3', a single prompt was used for both English and Swedish datasets. The speed of the model in generating Swedish responses was exceedingly slow. After seven hours, it only produced five outputs.

### 3.1.2. Translation Generation Tasks

In the appendix in Section A, in Figures 8 to 10, you will find a comprehensive list of all prompts used for the different models. The details of our implementation and observation of the translation generation task are as follows:

In our experimentation with 'Llama 3', we opted not to use the 'guidance' framework because of its ineffective performance. 'Llama 3' showed promising results for each language pair. We experimented with two different prompts, as shown in Figure 10 and observed instances where 'Llama 3' successfully generated hypotheses in the desired target language but struggled with the source language. Examples illustrating this phenomenon can be found in the Table 19.

Various prompts were tested, and the one that was chosen, as shown in Figure 9, showed effectiveness in generating the most automatic translations. However, 'GPT-3.5 Turbo' still struggled to instantly create translations (hyp- and hyp+) for all sources. The main issue was the variations in quotation marks which caused problems during the extraction process. As a result, we had to prompt some sentences individually (instead of being able to loop them as a group) so that the structure was recognized by GPT again. List 3.1.2 shows the number of samples had been done individually.

- German to English: with 12 sources where 3 sources needed to be translated manually.
- English to German: with 10 sources where 2 sources needed to be translated manually.
- French to English: with 19 sources where 3 sources needed to be translated manually.
- English to French: with 64 sources where 0 sources needed to be translated manually.

Translating from English was a smoother process for the Gemma model compared to translating to English.

### 3.2. Detection task

The detection task involves trial and test sets for both scenarios.

**Table 1**
Paraphrasing detection dataset details.

| Dataset | Type | Count |
|---|---|---|
| Eloquent/HalluciGen-PG | trial_detection_english | 15 |
| Eloquent/HalluciGen-PG | trial_detection_swedish | 19 |
| Eloquent/HalluciGen-PG | test_detection_english | 118 |
| Eloquent/HalluciGen-PG | test_detection_swedish | 118 |

### 3.2.1. Paraphrasing Detection Task

Table 1 shows the number of samples for each trial and test set for the paraphrasing detection task. The trial dataset for the paraphrasing detection task is structured as follows:

- **id:** Unique identifier of the example.
- **source:** Original model input for paraphrase generation.
- **hyp1:** First alternative paraphrase of the source.
- **hyp2:** Second alternative paraphrase of the source.
- **label:** *hyp1* or *hyp2*, based on which of those has been annotated as hallucination.
- **type:** Hallucination category assigned. Possible values include:
    - addition
    - named-entity
    - number
    - conversion
    - date
    - tense
    - negation
    - gender
    - pronoun
    - antonym
    - natural

We compared the performance of different models on the trial dataset using distinct prompts. Some prompts used for the paraphrasing detection task on the trial dataset are presented in Figure 11. Additionally, Tables 20 to 32 illustrate the performance of various prompts on the trial dataset for both English and Swedish.

A challenge with Gemma was its tendency to generate code within responses. We implemented a specific 'JSON' format to ensure retrievable output. Figure 12 indicates the example of generated output from Gemma. Figures 13 to 17 display the prompts employed in the paraphrasing detection task across various models for the test set.

### 3.2.2. Translation Detection Task

The following details are provided about the translation detection dataset.

- Both trial and test datasets include data for four language pairs as follows:
    - **de-en:** Source language: German, Target language: English
    - **en-de:** Source language: English, Target language: German
    - **fr-en:** Source language: French, Target language: English
    - **en-fr:** Source language: English, Target language: French

- The trial dataset included 10 data entries, with 5 entries featuring hallucination as hyp1 and the other 5 as hyp2. The structure of the trial dataset is illustrated below:
    - **id:** Unique identifier of the example.
    - **langpair:** Language of source and hypotheses pair
    - **source:** Source Text
    - **hyp1:** First alternative translation of the source.
    - **hyp2** Second alternative translation of the source.
    - **type:** Hallucination category assigned. Possible values include:
        * addition
        * named-entity
        * number
        * conversion
        * date
        * tense
        * negation
        * gender
        * pronoun
        * antonym
        * natural
    - **label** *hyp1* or *hyp2*, based on which of those has been annotated as hallucination
- In the test collection, there are 100 data samples for each language pair.
  The structure of the test dataset is presented as follows:
    - **id:** Unique identifier of the example.
    - **langpair:** Language of source and hypotheses pair
    - **source:** Source Text
    - **hyp1:** First alternative translation of the source.
    - **hyp2** Second alternative translation of the source.

Our implementation and observations of the translation detection task are delineated below, categorized according to each model.

**Observations for Llama 3** Ultimately, we experimented with 15 different prompts for the 'Llama 3' model. Among these, the prompt, as shown in Figure 18(a) yielded the most favorable results. Table 33 demonstrates the achieved results by using this prompt on the trial dataset. So, we opted for it for the final detection task.

The main observations for Llama 3 are:

- 'Llama 3' is not able to detect a label for every data entry (support is only 4 for each, hyp1 and hyp2). Figure 18 demonstrates the prompts used by 'Llama 3' on the test set.
- When detecting the hallucination, 'Llama 3' gives explanations, such as: 'I chose hyp1 as the hallucination because it contains a date (December 5) that is not present in the source text. The source text only mentions the date August 5, but hyp1 provides a different date.' The first row in Table 34 shows this issue.
- As both examples in Table 34 indicate 'Llama 3' exhibits gender bias. In the first example, it failed to recognize the feminine noun 'Wirtschaftsprüferin' shows a female auditor and labeled it as gender-neutral. It made a gender assumption in hyp1 which assumes a male auditor. Similarly, in the second example, 'Llama 3' struggled to understand the clear indication of a female secretary with the word 'Sekretärin.'

- 'Llama 3' struggles with understanding and converting measurements and it could not recognize when different units are essentially the same. For instance, it sees 'kilometers' and thinks it is different from 'metres' which leads to mistakenly identifying that text as a hallucination. Additionally, 'Llama 3' makes the assumption that hyp2 is the hallucination because it contains 'kilometers' instead of 'km' and it fails to consider the fact that hyp1 also uses 'metres' instead of 'km.' Table 35 highlights this issue.
- 'Llama 3' struggles to recognize the different ways dates can be written. As shown in Table 36, it could not understand that '21. Januar' and 'Jan. 21st' refer to the same date.
- In the end, we noticed that the prompt immensely influences the outcome of 'Llama 3'. When using different prompts, 'Llama 3' was either able to detect the gender, conversion, or the correct date, or it was not. For example, $row1$ in Table 37 shows that using the prompt shown in Figure 18(b), 'Llama 3' correctly explains that 'Wirtschaftsprüferin' refers to a female auditor in the first example, but then it mistakenly swaps hyp1 and hyp2. Additionally, as shown in the second row of this table, the new prompt allows 'Llama 3' to detect the correct gender indicated in the source text. However, 'Llama 3' still fails to assign the correct label. In the third row, we can see that 'Llama 3' correctly converts 65 km to 65,000 meters and identifies the hallucination in hyp2. Additionally, 'Llama 3' correctly identifies the wrong date in the last example. The primary issue with this prompt is that 'Llama 3' frequently fails to identify any hallucinations in certain data samples.

**Observations for GPT-3.5 Turbo and GPT-4** The approach used for 'GPT-3.5 Turbo' was replicated for 'GPT-4' to directly assess comprehension. Various prompts were tested, and two were selected based on the best results from previous trials.

The main observations for GPT-3.5 Turbo and GPT-4 are:

- There were some samples where no hallucinations were detected. Table 38 displays the count of failed examples for 'GPT-4' and 'GPT-3.5 Turbo' in the translation detection task. Additionally, Table 39 lists some samples for which 'GPT-4' failed to assign labels with our explanation for each one.
- Regarding the prompts for GPT models, both seem to encounter issues with misinterpretations or slightly inaccurate translations. Additionally, both struggle to identify incorrect pronouns.
- Initially, during the phase with incorrect trial datasets, it was observed that 'GPT-3.5 Turbo' had difficulty recognizing hallucinations when names were slightly misspelled or had an extra letter appended.

**Observations for Gemma** We tried various prompts, but Gemma showed better (80% Accuracy) in detecting the correct label when it was first asked to translate the hypothesis into the language of the source and then detect hallucinations. Figure 19 indicates the prompts used by Gemma on the test set.

The main observations for Gemma are:

- The performance was significantly worse when the prompts were too scientific or contained too many technical terms.
- Tricky samples for Gemma in the detection task include detecting the gender in comparison to the source (female/male), and identifying when numbers are incorrect, such as missing zeros.

**Observations for ensemble voting approach** We opted for a straightforward voting approach to ensemble model predictions due to the limitations imposed by the small sample size of the trial set. This method ensured all models contributed equally.

Since we compared an even number of models, there were instances where two models voted for hyp1 and the other two voted for hyp2. In these cases, we randomly selected the label.

**Table 2**
Test Dataset for Paraphrasing Detection Task Cross Model Evaluation.

| Dataset | Type | Count |
| --- | --- | --- |
| Eloquent/HalluciGen-PG | cross_model_evaluation_english | 594 |
| Eloquent/HalluciGen-PG | cross_model_evaluation_swedish | 380 |

### 3.3. Cross-evaluation task

The following provides detailed information regarding the cross-evaluation task. Table 2 presents information regarding the samples included in the paraphrasing task. The prompt showns in Figure 20 has been used for the english paraphrasing task.

In the translation task, sometimes none of the models detected any hallucinations in either hypothesis which results in some blank spaces in the CSV file due to the lack of predictions. Additionally, there were instances where no hallucinations were present because both hypotheses, hyp1 and hyp2, were the same.

## 4. Results

This part presents the results in detail for each task and scenario. It is worth noting that prior to showing the results from LLMs, Logistic Regression and Random Forest classifiers were used for an initial evaluation to establish a baseline performance for comparison with LLMs. Both LR and RF classifiers achieved similar performance with an F1-score of 0.5.

For the evaluation of the generation task, the lab employed a zero-shot text classification Natural language inference (NLI) model ('$MoritzLaurer/bge-m3-zeroshot-v2.0$') to predict whether 'hyp+' is entailed within the source sentence and whether 'hyp-' contradicts the source sentence. They used only two labels: 'entailment' and 'not_entailment'. This approach helps us assess whether the systems can produce coherent hyp+/hyp- pairs. It is important to note that the performance of the classification model is not perfect, but it demonstrated reasonable performance on the detection test set across various languages and language pairs [2].

For evaluating both detection and cross-model tasks, the lab reported key metrics such as Accuracy, F1-score, Precision, and Recall for each model. Additionally, several baseline models were evaluated by the lab. For cross-model assessment, the lab also employed two metrics: Matthews Correlation Coefficient (MCC) and Cohen's Kappa.

The Average MCC ($\overline{\text{MCC}}$) measures the quality of binary classifications by considering true and false positives and negatives, while the Standard Deviation of MCC ($\sigma_{MCC}$) provides insight into the consistency of the model's performance. Similarly, the Average Kappa ($\bar{\kappa}$) measures inter-rater reliability for categorical items, and the Standard Deviation of Kappa ($\sigma_\kappa$) indicates the variability or consistency of the Kappa metric [15].

Tables 3 to 5 demonstrate the evaluation of detection, generation, and cross-model evaluation for English paraphrasing tasks.

The performance of detection across various models on the English paraphrasing task is presented in Table 3. The model 'GPT-4' with prompt 'En_Se_Para_Det_GPT3.5_GPT4_v2' achieved the highest performance with Accuracy, F1-score, Precision, and Recall scores of 0.91.

Table 4 presents the results for the generation step. The model 'GPT-3.5 Turbo' with prompt 'En_Para_Gen_GPT3.5' achieved the highest performance in hyp+ entailment mean (0.964) and hyp+ correct label mean (0.983). Furthermore, The model 'Llama 3' with prompt 'En_Para_Gen_Llama3' showed strong performance in hyp- not entailment mean(0.978) and hyp- correct label mean (0.983).

Table 5 presents the results for the cross model. The model 'GPT-4' with prompt 'final_gpt4_en_v2_cross_model_detection' showed Accuracy, F1-score, Precision, and

Recall scores of 0.93. In the next stage, the majority model with prompt 'majority_vote_cross_model_result_en' demonstrated impressive performance.

Table 6 shows that the model with prompt 'majority_vote_cross_model_result_en' achieved the highest performance with an average MCC of 0.83 and average Kappa of 0.81.

Table 7 presents the performance metrics of various models in the detection step for the Swedish paraphrasing task. The model 'GPT-4' with prompt 'En_Se_Para_Det_GPT3.5_GPT4_v1 (GPT4)' achieved an Accuracy score of 0.81 and with consistent scores across all metrics (F1 = 0.81, Precision = 0.81, Recall = 0.81). Additionally, the baseline 'baseline-bge-m3-zeroshot-v2.0/sv_bge-m3-zeroshot-v2.0' shows the highest Accuracy of 0.92 across all models.

Table 8 summarizes the results of models in the generation step for Swedish paraphrasing where the focus is on metrics related to hypothesis entailment and not_entailment. The model 'GPT-3.5 Turbo' with prompt 'Se_Para_Gen_GPT3.5' demonstrated strong performance with high scores in hyp+ entailment mean of 0.88, hyp+ correct label mean of 0.90, hyp- contradiction mean of 0.91, and hyp- correct label mean of 0.93.

Tables 9 and 10 present cross-model evaluation results for the Swedish paraphrasing task that highlights the model performance across different evaluation criteria. The model majority voting with prompt 'majority_vote_cross_model_result_se' showed competitive performance. Table 10 provides statistical measures for models excluding baselines that indicate the noted majority voting technique has consistency and reliability.

Tables 11 to 14 report the performance of English-French translation detection, generation, and cross-model evaluation.

Table 11 highlights several key points regarding the performance of the detection task. Model 'GPT-4' with prompt 'results_gpt4_en_fr' achieved the highest performance with Accuracy, F1-score, and Recall of 0.90, and Precision of 0.91. Additionally, we can observe that the majority voting model with prompt 'majority_vote_result_en_fr' also performed well with Accuracy, F1-score, and Recall of 0.83 and Precision of 0.86.

One of the conclusions can be drawn from Table 12 is that the baseline model 'baseline-general-prompt/en-fr.gen' showed a better performance with hyp+ entailment mean 0.90 and hyp+ correct label mean of 0.93, while it has a lower performance in hyp- contradiction mean of 0.10 and hyp- correct label mean of 0.08. Also, it is clear that model 'GPT-3.5 Turbo' with prompt 'results_gpt_en_fr' demonstrated a high performance in hyp- contradiction mean of 0.88, and hyp- correct label mean of 0.91.

From tables 13 and 14 we have the finding that the majority voting approach with prompt 'majority_vote_result_en_fr' reached Accuracy 0.79, F1 score 0.78, Precision 0.80, and Recall 0.79. This combination exhibited the highest average MCC 0.66 and average Kappa 0.65.

Tables 15 to 18 report the results for the evaluation of the English-German translation detection, generation, and cross-model.

The important observation from the Table 15 is that the model 'GPT-4' along with prompt 'results_gpt4_en_de' showed the highest performance with an Accuracy, F1 score, and Recall all at 0.86 and Precision 0.89.

From Table 16 we can see that the model 'GPT-3.5 Turbo' with the mixture of prompt 'results_gpt_en_de' exhibited better performance in hyp- contradiction mean of 0.83, and hyp- correct label mean of 0.84. 'Gemma' with prompt 'En_De_Trans_Gen_gamma' showed the best hyp+ correct label mean of 0.85. Additionally, 'baseline-phenomena-mentions-prompt/en-de.gen' provides a better hyp+ entailment mean of 0.84.

Tables 17 and 18 provide the insight that the model 'GPT-3.5 Turbo' with 'results_gpt_en_de' had the highest Accuracy of 0.76, F1 score of 0.75, Precision of 0.77, and Recall of 0.76. The prompt 'majority_vote_result_en_de' for majority voting had the highest average MCC of 0.60 and average Kappa of 0.58 which indicates strong inter-model agreement and consistency.

**Table 3**
Results of the **English** Detection Step Paraphrasing Task

| Model | Accuracy | F1 | Precision | Recall |
|---|---|---|---|---|
| En_Se_Para_Det_Llama3_v2 | 0.69 | 0.69 | 0.81 | 0.69 |
| En_Se_Para_Det_GPT3.5_GPT4_v2 | 0.91 | 0.91 | 0.91 | 0.91 |
| En_Para_Det_Llama3_v1 | 0.80 | 0.80 | 0.81 | 0.80 |
| En_Para_Det_Gemma_v1 | 0.71 | 0.71 | 0.77 | 0.71 |
| En_Se_Para_Det_GPT3.5_GPT4_v1 (GPT4) | 0.73 | 0.73 | 0.83 | 0.73 |
| En_Se_Para_Det_Gemma_v2 | 0.54 | 0.49 | 0.73 | 0.54 |
| majority_vote_result_en_prompt (Prompts of Version 2) | 0.85 | 0.85 | 0.86 | 0.85 |
| En_Se_Para_Det_GPT3.5_GPT4_v1 (GPT3.5) | 0.68 | 0.68 | 0.75 | 0.68 |
| **Baseline Models** | | | | |
| baseline-bge-m3-zeroshot-v2.0/en_bge-m3-zeroshot-v2.0 | 0.90 | 0.90 | 0.90 | 0.90 |
| baseline-llama2-meaning-detection/en.det | 0.45 | 0.44 | 0.44 | 0.45 |
| baseline-llama2-not-supported-detection/en.det | 0.34 | 0.35 | 0.39 | 0.34 |
| baseline-llama2-paraphrase-detection/en.det | 0.34 | 0.35 | 0.37 | 0.34 |

**Table 4**
Results of the **English** Generation Step for Paraphrasing Task

| Model | hyp+ entailment mean | hyp+ correct label mean | hyp- contradiction mean | hyp- correct label mean |
|---|---|---|---|---|
| En_Para_Gen_Gemma_v2 | 0.828 | 0.857 | 0.894 | 0.908 |
| En_Para_Gen_Gemma_v1 | 0.782 | 0.824 | 0.894 | 0.899 |
| En_Para_Gen_GPT3.5 | 0.964 | 0.983 | 0.797 | 0.807 |
| En_Para_Gen_Llama3 | 0.843 | 0.882 | 0.978 | 0.983 |
| **Baseline Models** | | | | |
| baseline-mixtral-8x7b-instruct-hallucination-detection/en.gen | 0.920 | 0.924 | 0.738 | 0.748 |

**Table 5**
Results of the **English** Cross-Model Evaluation for Paraphrasing Task

| Model | Accuracy | F1 | Precision | Recall |
|---|---|---|---|---|
| final_gpt35_en_v2_cross_model_detection | 0.88 | 0.88 | 0.89 | 0.88 |
| majority_vote_cross_model_result_en | 0.92 | 0.92 | 0.93 | 0.92 |
| final_lama3_cross_model_en_v1 | 0.87 | 0.87 | 0.88 | 0.87 |
| final_gpt4_en_v2_cross_model_detection | 0.93 | 0.93 | 0.93 | 0.93 |
| final_gemma_en_v1_cross_model | 0.78 | 0.77 | 0.83 | 0.78 |
| **Baseline Models** | | | | |
| baseline-bge-m3-zeroshot-v2.0/en.det.csv | 0.95 | 0.95 | 0.95 | 0.95 |

**Table 6**
Results of the **English** Cross-Model Evaluation (excluding the baseline models) for Paraphrasing Task

| Model | $\overline{\text{MCC}}$ | $\sigma_{MCC}$ | $\bar{\kappa}$ | $\sigma_{\kappa}$ |
|---|---|---|---|---|
| final_gemma_en_v1_cross_model | 0.65 | 0.03 | 0.61 | 0.04 |
| final_gpt35_en_v2_cross_model_detection | 0.77 | 0.08 | 0.77 | 0.09 |
| final_gpt4_en_v2_cross_model_detection | 0.76 | 0.10 | 0.75 | 0.12 |
| final_lama3_cross_model_en_v1 | 0.75 | 0.09 | 0.74 | 0.10 |
| majority_vote_cross_model_result_en | 0.83 | 0.09 | 0.81 | 0.10 |

**Table 7**
Results of the **Swedish** Detection Step for Paraphrasing Task

| Model | Accuracy | F1 | Precision | Recall |
|---|---|---|---|---|
| En_Se_Para_Det_GPT3.5_GPT4_v1 (GPT4) | 0.81 | 0.81 | 0.81 | 0.81 |
| Se_Para_Det_Gemma_v1 | 0.59 | 0.52 | 0.71 | 0.59 |
| majority_vote_result_se (Prompts from Version 1) | 0.67 | 0.66 | 0.72 | 0.67 |
| En_Se_Para_Det_Gemma_v2 | 0.07 | 0.11 | 0.47 | 0.07 |
| En_Se_Para_Det_GPT3.5_GPT4_v1 (GPT 3.5) | 0.71 | 0.70 | 0.76 | 0.71 |
| En_Se_Para_Det_GPT3.5_GPT4_v2 (GPT 3.5) | 0.61 | 0.60 | 0.65 | 0.61 |
| En_Se_Para_Det_Llama3_v2 | 0.57 | 0.48 | 0.77 | 0.57 |
| Se_Para_Det_Llama3_v1 | 0.60 | 0.59 | 0.60 | 0.60 |
| **Baseline Models** | | | | |
| baseline-bge-m3-zeroshot-v2.0/sv_bge-m3-zeroshot-v2.0 | 0.92 | 0.92 | 0.92 | 0.92 |
| baseline-llama2-meaning-detection/sv.det | 0.60 | 0.60 | 0.62 | 0.60 |
| baseline-llama2-not-supported-detection/sv.det | 0.57 | 0.56 | 0.70 | 0.57 |
| baseline-llama2-paraphrase-detection/sv.det | 0.61 | 0.59 | 0.68 | 0.61 |
| sv_scandi-nli-large | 0.92 | 0.92 | 0.92 | 0.92 |

**Table 8**
Results of the **Swedish** Generation Step for Paraphrasing Task

| Model | hyp+ entailment mean | hyp+ correct label mean | hyp- contradiction mean | hyp- correct label mean |
|---|---|---|---|---|
| Se_Para_Gen_Gemma_v1 | 0.346 | 0.355 | 0.931 | 0.934 |
| Se_Para_Gen_Gemma_v2 | 0.588 | 0.618 | 0.710 | 0.697 |
| Se_Para_Gen_GPT3.5 | 0.881 | 0.908 | 0.918 | 0.934 |
| **Baseline Models** | | | | |
| baseline-gpt-sw3-6.7b-v2-hallucination-detection/sv.gen | 0.637 | 0.645 | 0.502 | 0.500 |
| baseline-mixtral-8x7b-instruct-hallucination-detection/sv.gen | 0.809 | 0.842 | 0.386 | 0.355 |

**Table 9**
Results of the **Swedish** Cross-Model Evaluation for Paraphrasing Task

| Model | Accuracy | F1 | Precision | Recall |
|---|---|---|---|---|
| final_lama3_cross_model_se_v1 | 0.71 | 0.70 | 0.71 | 0.71 |
| final_gpt35_se_v2_cross_model_detection | 0.68 | 0.68 | 0.69 | 0.68 |
| majority_vote_cross_model_result_se | 0.76 | 0.76 | 0.77 | 0.76 |
| final_gpt4_se_v2_cross_model_detection | 0.72 | 0.74 | 0.77 | 0.72 |
| final_gemma_se_v1_cross_model | 0.56 | 0.48 | 0.63 | 0.56 |
| **Baseline Models** | | | | |
| baseline-bge-m3-zeroshot-v2.0/sv.det | 0.75 | 0.75 | 0.75 | 0.75 |

**Table 10**
Results of the **Swedish** Cross-Model Evaluation (excluding the baseline models)

| Model | $\overline{\text{MCC}}$ | $\sigma_{MCC}$ | $\bar{\kappa}$ | $\sigma_{\kappa}$ |
|---|---|---|---|---|
| final_gemma_se_v1_cross_model | 0.26 | 0.04 | 0.19 | 0.04 |
| final_gpt35_se_v2_cross_model_detection | 0.51 | 0.18 | 0.48 | 0.20 |
| final_gpt4_se_v2_cross_model_detection | 0.46 | 0.16 | 0.41 | 0.19 |
| final_lama3_cross_model_se_v1 | 0.52 | 0.22 | 0.50 | 0.24 |
| majority_vote_cross_model_result_se | 0.62 | 0.19 | 0.59 | 0.23 |

**Table 11**
Results of the **English-French Translation** Detection Step

| Model | Accuracy | F1 | Precision | Recall |
|---|---|---|---|---|
| results_gpt4_en_fr_prompt2 | 0.79 | 0.79 | 0.79 | 0.79 |
| En_Fr_Trans_Det_llama3_v2 | 0.66 | 0.65 | 0.74 | 0.66 |
| En_Fr_Trans_Det_gemma_v1 | 0.66 | 0.66 | 0.67 | 0.66 |
| results_gpt_en_fr | 0.74 | 0.74 | 0.81 | 0.74 |
| En_Fr_Trans_Det_gemma_v2 | 0.63 | 0.60 | 0.81 | 0.63 |
| En_Fr_Trans_Det_llama3_v1 | 0.56 | 0.51 | 0.79 | 0.56 |
| results_gpt_en_fr_prompt2 | 0.76 | 0.76 | 0.83 | 0.76 |
| majority_vote_result_en_fr | 0.83 | 0.83 | 0.86 | 0.83 |
| results_gpt4_en_fr | 0.90 | 0.90 | 0.91 | 0.90 |
| **Baseline Models** | | | | |
| en_fr_bge-m3-zeroshot-v2.0 | 0.82 | 0.82 | 0.82 | 0.82 |
| baseline-general-detection-prompt_en-fr.det | 0.47 | 0.47 | 0.52 | 0.47 |
| baseline-meaning-detection-prompt_en-fr.det | 0.49 | 0.50 | 0.50 | 0.49 |
| baseline-supported-detection-prompt_en-fr.det | 0.40 | 0.24 | 0.17 | 0.40 |

**Table 12**
Results of the **English-French Translation** Generation Step

| Model | hyp+ entailment mean | hyp+ correct label mean | hyp- contradiction mean | hyp- correct label mean |
|---|---|---|---|---|
| En_Fr_Trans_Gen_llama3_v2 | 0.80277 | 0.81 | 0.82427 | 0.86 |
| En_Fr_Trans_Gen_gemma | 0.80958 | 0.8 | 0.50219 | 0.49 |
| results_gpt_en_fr | 0.85611 | 0.88 | 0.88556 | 0.91 |
| En_Fr_Trans_Gen_llama3_v1 | 0.75679 | 0.77 | 0.78892 | 0.81 |
| **Baseline Models** | | | | |
| baseline-phenomena-mentions-prompt/en-fr.gen | 0.89575 | 0.92 | 0.26324 | 0.23 |
| baseline-general-prompt/en-fr.gen | 0.90503 | 0.93 | 0.10912 | 0.08 |

**Table 13**
Results of the **English-French Translation** Cross-Model Evaluation

| Model | Accuracy | F1 | Precision | Recall |
|---|---|---|---|---|
| results_gpt_en_fr | 0.77 | 0.77 | 0.79 | 0.77 |
| results_gemma_en_fr_final | 0.57 | 0.57 | 0.57 | 0.57 |
| results_llama3_en_fr_final | 0.68 | 0.65 | 0.78 | 0.68 |
| majority_vote_result_en_fr | 0.79 | 0.78 | 0.80 | 0.79 |
| results_gpt4_en_fr | 0.77 | 0.76 | 0.79 | 0.77 |
| **Baseline Models** | | | | |
| baseline-general-detection-prompt/en-fr.cme | 0.44 | 0.45 | 0.46 | 0.44 |
| baseline-meaning-detection-prompt/en-fr.cme | 0.47 | 0.47 | 0.48 | 0.47 |
| baseline-supported-detection-prompt/en-fr.cme | 0.48 | 0.32 | 0.24 | 0.48 |

**Table 14**
Results of the **English-French Translation** Cross-Model Evaluation (excluding the baseline models)

| Model | $\overline{\text{MCC}}$ | $\sigma_{MCC}$ | $\bar{\kappa}$ | $\sigma_{\kappa}$ |
|---|---|---|---|---|
| majority_vote_result_en_fr | 0.66 | 0.23 | 0.65 | 0.24 |
| results_gemma_en_fr_final | 0.25 | 0.04 | 0.23 | 0.05 |
| results_gpt4_en_fr | 0.60 | 0.26 | 0.59 | 0.27 |
| results_gpt_en_fr | 0.59 | 0.26 | 0.57 | 0.27 |
| results_llama3_en_fr_final | 0.48 | 0.16 | 0.43 | 0.16 |

**Table 15**
Results of the **English-German Translation** Detection Step

| Model | Accuracy | F1 | Precision | Recall |
|---|---|---|---|---|
| majority_vote_result_en_de | 0.81 | 0.81 | 0.87 | 0.81 |
| En_De_Trans_Det_gemma_v1 | 0.59 | 0.59 | 0.60 | 0.59 |
| results_gpt4_en_de_prompt2 | 0.79 | 0.79 | 0.81 | 0.79 |
| En_De_Trans_Det_llama3_v1 | 0.54 | 0.47 | 0.78 | 0.54 |
| En_De_Trans_Det_llama3_v2 | 0.70 | 0.69 | 0.79 | 0.70 |
| En_De_Trans_Det_gemma_v2 | 0.58 | 0.54 | 0.73 | 0.58 |
| results_gpt_en_de_prompt2 | 0.80 | 0.80 | 0.85 | 0.80 |
| results_gpt4_en_de | 0.86 | 0.86 | 0.89 | 0.86 |
| results_gpt_en_de_prompt2 (duplicate?) | 0.68 | 0.67 | 0.77 | 0.68 |
| **Baseline Models** | | | | |
| en_de_bge-m3-zeroshot-v2.0 | 0.73 | 0.73 | 0.74 | 0.73 |
| baseline-general-detection-prompt/en-de.det | 0.48 | 0.48 | 0.51 | 0.48 |
| baseline-meaning-detection-prompt/en-de.det | 0.36 | 0.36 | 0.36 | 0.36 |
| baseline-supported-detection-prompt/en-de.det | 0.41 | 0.25 | 0.18 | 0.41 |

**Table 16**
Results of the **English-German Translation** Generation Step

| Model | hyp+ entailment mean | hyp+ correct label mean | hyp- contradiction mean | hyp- correct label mean |
|---|---|---|---|---|
| results_gpt_en_de | 0.75706 | 0.81 | 0.83454 | 0.84 |
| En_De_Trans_Gen_gamma | 0.82415 | 0.85 | 0.44845 | 0.42 |
| En_De_Trans_Gen_llama3_v2 | 0.81966 | 0.84 | 0.64852 | 0.68 |
| En_De_Trans_Gen_llama3_v1 | 0.78468 | 0.84 | 0.81466 | 0.84 |
| **Baseline Models** | | | | |
| baseline-phenomena-mentions-prompt/en-de.gen | 0.8455 | 0.85 | 0.35101 | 0.33 |
| baseline-general-prompt/en-de.gen | 0.83701 | 0.85 | 0.20566 | 0.19 |

**Table 17**
Results of the **English-German Translation** Cross-Model Evaluation

| Model | Accuracy | F1 | Precision | Recall |
|---|---|---|---|---|
| majority_vote_result_en_de | 0.75 | 0.74 | 0.76 | 0.75 |
| results_llama3_en_de_final | 0.58 | 0.52 | 0.69 | 0.58 |
| results_gemma_en_de_final | 0.53 | 0.53 | 0.53 | 0.53 |
| results_gpt4_en_de | 0.73 | 0.73 | 0.76 | 0.73 |
| results_gpt_en_de | 0.76 | 0.75 | 0.77 | 0.76 |
| **Baseline Models** | | | | |
| baseline-general-detection-prompt/en-de.cme | 0.42 | 0.43 | 0.45 | 0.42 |
| baseline-meaning-detection-prompt/en-de.cme | 0.41 | 0.42 | 0.43 | 0.41 |
| baseline-supported-detection-prompt/en-de.cme | 0.49 | 0.33 | 0.58 | 0.49 |

**Table 18**
Results of the **English-German Translation** Cross-Model Evaluation (excluding the baseline models)

| Model | $\overline{\text{MCC}}$ | $\sigma_{MCC}$ | $\bar{\kappa}$ | $\sigma_{\kappa}$ |
|---|---|---|---|---|
| majority_vote_result_en_de | 0.60 | 0.28 | 0.58 | 0.30 |
| results_gemma_en_de_final | 0.16 | 0.05 | 0.15 | 0.05 |
| results_gpt4_en_de | 0.53 | 0.31 | 0.52 | 0.33 |
| results_gpt_en_de | 0.52 | 0.32 | 0.50 | 0.33 |
| results_llama3_en_de_final | 0.34 | 0.11 | 0.27 | 0.10 |

# 5. Conclusion

In conclusion, this study leveraged several LLMs to investigate both the generation and detection of hallucinations by LLMs themselves. The four distinct models employed presented their own unique evaluation challenges. We explored various prompt techniques including few-shot learning and chain of thought by using the guidance framework. Additionally, for the detection task, we tested an ensemble voting approach to combine the results from different LLMs. Although in this study we could achieve better results in comparison to the baseline models, our findings indicate that while some issues can be addressed through effective prompting, others remain difficult to mitigate solely by prompt engineering. Moreover, identifying the optimal prompt itself poses a significant challenge.

## A. Appendix

```
1  answer_format = {"hyp+": "", "hyp-": ""}
2  user_prompt = f'''
3  <start\_of\_turn>user
4  Given the src below, generate a paraphrase
        hypothesis hyp+ that is supported by
        src and a second paraphrase hyp- that
        is not supported by src.
5  Provide the result in the following format:
         {answer_format}
6  Src: {source}
7  <end\_of\_turn>
8  Result:
9  <start\_of\_turn>model
10
```
**Listing 1:** En_Para_Gen _Gemma_v1

```
1  answer_format = {"hyp+": "", "hyp-": ""}
2  user_prompt = f'''
3  <start\_of\_turn>
4  Med tanke på källan nedan, generera en
        parafras-hypotes hyp+ som stöds av kä
        llan och en andra parafras hyp- som
        inte stöds av källan.
5  Ge resultatet i följande format: {
        answer_format}
6  Källa: {source}
7  <end\_of\_turn>
8  Resultat:
9  <start\_of\_turn>modell
10
```
**Listing 2:** Se_Para_Gen_Gemma_v1

**Figure 4:** First Prompts used by 'Gemma' for Paraphrasing Generation Task.

```
1       Given the src below, generate a
    paraphrase hypothesis  hyp+ that is
    supported by src and a second
    paraphrase hyp- that is not supported
    by src.
2       Provide the result in the following
     format: {"hyp+": "", "hyp-": ""}
3       Src: {source}
4       Result:
5
```
**Listing 3:** En_Para_Gen _GPT3.5

```
1          answer_format = {"hyp+": " ", "hyp-
    ": " "}
2      user_prompt = f'''
3      Med tanke på källan nedan, generera en
     parafras-hypotes hyp+ som stöds av kä
    llan och en andra parafras hyp- som
    inte stöds av källan.
4      Ge resultatet i följande format: {
    answer_format}
5      Källa: {source}
6      Resultat:
7      '''
8   with system():
9       lm = gpt + "Du är en textgenerator.
     Du är specialiserad på att
    parafrasera texter"
10  with user():
11      lm += user_prompt
12  with assistant():
13      lm += gen("answer")
14  result = lm["answer"]
15
```
**Listing 4:** Se_Para_Gen_GPT3.5

**Figure 5:** Prompts used by 'GPT-3.5 Turbo' for Paraphrasing Generation Task.

```
1           Given the src below, generate a paraphrase hypothesis  hyp+ that is
    supported by src and a second paraphrase hyp- that is not supported by src.
2
3   Provide the result in the following format: {"hyp+": "", "hyp-": ""}
4   Src: {source}
5   Result: }
6       Result:
7
```

**Listing 5:** En_Para_Gen _Llama3

**Figure 6:** Prompt used by 'Llama 3' for Paraphrasing Generation Task.

**Table 19**

LLama3 generates hyp+ and hyp- in the source language instead of the desired target language.

| Lang-pair | Source | hyp+ | hyp- |
|---|---|---|---|
| en-de | All nouns, alongside the word Sie for you, always begin with a capital letter, even in the middle of a sentence. | All nouns, alongside the word Sie for you, always begin with a capital letter, even in the middle of a sentence, except for those that are part of a title or a proper noun. | All nouns, alongside the word sie for you, always begin with a capital letter, even in the middle of a sentence. |
| en-fr | The final line of the third verse was changed during the reign of Alexander I of Yugoslavia in "Kralja Aleksandra, Bože hrani ". | The final line of the third verse was modified during the reign of Alexander I of Yugoslavia in 'Kralja Aleksandra, Bože hrani'. | The final line of the third verse was rewritten during the reign of Alexander the Great in 'Kralja Aleksandra, Bože hrani'. |

**Table 20**

Comparison of English Prompt 1 Results on Trial Dataset from Gemma, Llama3, and GPT-3.5 for the paraphrasing detection task.

| id | type | label | prediction gemma | prediction gpt 3.5 | prediction llama3 |
|---|---|---|---|---|---|
| 0 | antonym | hyp1 | hyp1 | hyp1 | hyp1 |
| 1 | negation | hyp1 | hyp1 | hyp1 | hyp1 |
| 2 | antonym | hyp1 | hyp1 | hyp1 | hyp1 |
| 3 | named entity | hyp1 | hyp1 | hyp1 | hyp1 |
| 4 | natural | hyp2 | hyp2 | hyp2 | hyp2 |
| 5 | addition | hyp2 | hyp2 | hyp2 | hyp2 |
| 6 | gender | hyp2 | hyp2 | hyp2 | hyp2 |
| 7 | natural | hyp1 | hyp2 | hyp1 | hyp2 |
| 8 | number | hyp2 | hyp2 | hyp2 | hyp2 |
| 9 | pronoun | hyp1 | hyp2 | hyp2 | hyp2 |
| 10 | pronoun | hyp1 | hyp1 | hyp2 | hyp2 |
| 11 | addition | hyp2 | hyp2 | hyp2 | hyp2 |
| 12 | conversion | hyp1 | hyp2 | hyp1 | hyp1 |
| 13 | natural | hyp2 | hyp2 | hyp1 | hyp2 |
| 14 | named entity | hyp2 | hyp2 | hyp2 | hyp2 |
| 15 | date | hyp1 | hyp1 | hyp1 | hyp1 |

```
1  answer_format = {"hyp+": "", "hyp-": ""}
2  user_prompt = f'''
3  <start_of_turn>user
4  As an AI model, your task is to generate two paraphrases based on the given source text. The
       first paraphrase, labeled as 'hyp+', should be supported by the source text. The second
       paraphrase, labeled as 'hyp-', should not be supported by the source text.
5  Here's an example to illustrate this:
6  Source: The population has declined in some 210 of the 280 municipalities in Sweden, mainly
       in inland central and northern Sweden.
7  hyp+: In the majority of Sweden's 280 municipalities, the population has gone down.
8  This is a paraphrase that is supported by the source. It's saying essentially the same thing
       as the source: the population has decreased in most municipalities. The wording is
       different, but the meaning is the same. Hence, it's labeled as 'hyp+'.
9  hyp-: In the majority of Sweden's 280 municipalities, the population has gone up.
10 This is a paraphrase that is not supported by the source. It's saying the opposite of what
       the source says: the population has increased in most municipalities. This contradicts
       the information in the source. Hence, it's labeled as 'hyp-'.
11 Now, given the source text below, generate 'hyp+' and 'hyp-' paraphrases and provide the
       result in the following format: {answer_format}
12 Source: {source}
13 <end_of_turn>
14 Result:
15 <start_of_turn>model
16
```

**Listing 6:** En_Para_Gen _Gemma_v2

```
1  answer_format = {"hyp+": "", "hyp-": ""}
2  user_prompt = f'''
3  <start_of_turn>användare
4  Som en AI-modell är din uppgift att generera två parafraser baserade på den angivna
       källtexten. Den första parafrasen, märkt som 'hyp+', ska stödjas av källtexten. Den andra
       parafrasen, märkt som 'hyp-', ska inte stödjas av källtexten.
5  Här är ett exempel för att illustrera detta:
6  Källa: Intäkterna från mjukvarulicenser, ett mått som finansanalytiker följer noga, minskade
       med 21 procent till 107,6 miljoner dollar.
7  hyp+: Intäkter från programvarulicenser, en metrik som noggrant övervakas av finansiella
       analytiker, minskade med 21 procent till ett belopp av 107,6 miljoner dollar.
8  Detta är en parafras som stöds av källan. Den säger i princip samma sak som källan:
       intäkterna har minskat i de flesta kommunerna. Formuleringen är annorlunda, men
       betydelsen är densamma. Därför märks den som 'hyp+'.
9  Detta är en parafras som stöds av källan. Den säger i princip samma sak som källan:
       intäkterna har minskat i de flesta kommunerna. Formuleringen är annorlunda, men
       betydelsen är densamma. Därför märks den som 'hyp+'.
10 hyp-: Intäkter från programvarulicenser, en metrik som noggrant övervakas av finansiella
       analytiker, minskade med 42 procent till ett belopp av 107,6 miljoner dollar.
11 Detta är en parafras som inte stöds av källan. Den säger motsatsen till vad källan säger:
       intäkterna har ökat i de flesta kommunerna. Detta motsäger informationen i källan. Därför
       märks den som 'hyp-'.
12 Nu, med den angivna källtexten nedan, generera 'hyp+' och 'hyp-' parafraser och ge resultatet
       i följande format: {answer_format}
13 Källa: {source}
14 <end_of_turn>
15 Resultat:
16 <start_of_turn>modell
17
```

**Listing 7:** Se_Para_Gen_Gemma_v2

**Figure 7:** Second Prompts used by 'Gemma' for Paraphrasing Generation Task.

```
1 user_prompt = f'''
2 You are a text generator and your task is to generate two translation hypothesis given the 'src'
      below.
3 The first translation labelled as 'hyp+' should be supported by 'src' and the second
      translation labelled as 'hyp-' should not be supported by 'src'.
4 Provide the result in the following format: "hyp+": "", "hyp-": "". Target language: "English"
5
6 Src: {source}
7
8 Result:
9
```

**Listing 8:** En_De_Trans_Gen_gamma, En_Fr_Trans_Gen_gemma

**Figure 8:** Prompts used by 'Gemma' for Translation Generation Task.

```
1         answer_format = {"hyp+": "", "hyp-": ""}
2         system_msg = "You are a text generator for translation"
3         user_prompt = f'''
4         Your task is to generate two translation hypothesis given the 'src' below. The first
    translation labelled as 'hyp' should be supported by 'src' and the second translation
    labelled as 'hyp-' should not be supported by "src". Provide the result in the following
    format: {answer_format}. Target language: {target_language}
5
6         Src: {source}
7
8         Result:
9         '''
10
```

**Listing 9:** En_De_Trans_Gen_gamma, En_Fr_Trans_Gen_gemma

**Figure 9:** Prompts used by 'GPT-3.5 Turbo' and 'GPT-4' models for Translation Generation Task.

### Table 21
Classification Report Gemma of English Prompt 1 Results on Trial Dataset

|              | Precision | Recall | F1-Score | Support |
|--------------|-----------|--------|----------|---------|
| **hyp1**     | 1.00      | 0.78   | 0.88     | 9       |
| **hyp2**     | 0.78      | 1.00   | 0.88     | 7       |
| **Accuracy** |           |        | 0.88     | 16      |
| **Macro avg**    | 0.89  | 0.89   | 0.88     | 16      |
| **Weighted avg** | 0.90  | 0.88   | 0.88     | 16      |

### Table 22
Classification Report GPT-3.5 of English Prompt 1 Results on Trial Dataset.

|              | Precision | Recall | F1-Score | Support |
|--------------|-----------|--------|----------|---------|
| **hyp1**     | 1.00      | 0.67   | 0.80     | 9       |
| **hyp2**     | 0.70      | 1.00   | 0.82     | 7       |
| **Accuracy** |           |        | 0.81     | 16      |
| **Macro avg**    | 0.85  | 0.83   | 0.81     | 16      |
| **Weighted avg** | 0.87  | 0.81   | 0.81     | 16      |

```
1          answer_format = {"hyp+": "", "hyp-": ""}
2          user_prompt = f'''
3          Given the 'src' below, generate a translation hypothesis 'hyp+' that is supported by '
    src' and a second translation 'hyp-' that is not supported by 'src'.
4          Provide the result in the following format: {answer_format}.
5          Target language: {target_language}
6          Src: {source}
7          Result:
8
```

Listing 10: De_En_Trans_Gen_llama3_v1, En_De_Trans_Gen_llama3_v1, Fr_En_Trans_Gen_llama3_v1, En_Fr_Trans_Gen_llama3_v1

```
1          answer_format = {"hyp+": "", "hyp-": ""}
2          user_prompt = f'''
3          As an AI model, your task is to generate two translation hypothesis given the 'src'
    below. The first translation labelled as 'hyp+' should be supported by 'src' and the
    second translation labelled as 'hyp-' should not be supported by 'src'.
4          Provide the result in the following format: {answer_format}.
5          Target language: {target_language}
6          Here is an example to illustrate this:
7          src: The days in the summer can lead to problems getting sufficient sleep and
    associated health issues.
8          hyp+: Die Tage im Sommer können zu Problemen führen, genügend Schlaf zu bekommen und
    damit verbundene Gesundheitsprobleme.
9          This is a translation that is supported by the source. It is the exact translation of '
    src'.
10         hyp-: Die sehr langen Tage im Sommer können zu Problemen führen, genügend Schlaf zu
    bekommen und damit verbundene Gesundheitsprobleme.
11         This is a translation, which is not supported by the source. It includes an addition,
    which is "sehr langen" in this case. In 'src', there is no mention about 'the very long'
    days in the summer. Hence, it is labelled as 'hyp-'.
12         Src: {source}
13         Result:
14
```

Listing 11: En_De_Trans_Gen_llama3_v2, other language pairs contain an example in the associated language

**Figure 10:** Prompt used by 'Llama 3' for Translation Generation Task.

**Table 23**
Classification Report Llama3 of English Prompt 1 Results on Trial Dataset.

|              | Precision | Recall | F1-Score | Support |
|--------------|-----------|--------|----------|---------|
| **hyp1**     | 1.00      | 0.78   | 0.88     | 9       |
| **hyp2**     | 0.78      | 1.00   | 0.88     | 7       |
| **Accuracy** |           |        | 0.88     | 16      |
| **Macro avg**    | 0.89  | 0.89   | 0.88     | 16      |
| **Weighted avg** | 0.90  | 0.88   | 0.88     | 16      |

```
1            answer_format = {"label": ""}
2            user_prompt = f'''
3            <start_of_turn>user
4            Givet en "src" och två hypoteser "hyp1" och "hyp2" är din uppgift att upptäcka vilken av de två hypoteserna ("label") som inte stöds av källan.
5            Ge resultatet i följande format: {answer_format}.
6
7            Src: {source}
8            hyp1 : {hyp1}
9            hyp2 : {hyp2}
10
11           <end_of_turn>
12
13           Resultat:
14           <start_of_turn>model
15           '''
16
17
```

**Listing 12:** Swedish Prompt 1.

```
1            answer_format = {"label": ""}
2            user_prompt = f'''
3            <start_of_turn>user
4            You are a researcher investigating a new phenomenon. You have gathered data ({source}) and formulated two competing hypotheses (hyp1: {hyp1}, and hyp2: {hyp2}) to explain it.
5            Identify the hypothesis that contradicts the information provided in the given source.
6            Provide the result in the following format: {answer_format}.
7
8            <end_of_turn>
9
10           Result:
11           <start_of_turn>model
12           '''
13
```

**Listing 13:** English Prompt 2.

```
1            answer_format = {"label": ""}
2            user_prompt = f'''
3            <start_of_turn>user
4            Given a "src" and two hypotheses "hyp1" and "hyp2" your task is to detect which of the two hypotheses ("label") is not supported by the source.
5            Provide the result in the following format:
6            {answer_format}.
7
8            Src: {source}
9            hyp1: {hyp1}
10           hyp2: {hyp2}
11
12           <end_of_turn>
13
14           Result:
15           <start_of_turn>model
16           '''
17
```

**Listing 14:** English Prompt 1 and Swedish Prompt 2.

**Figure 11:** Prompts used for English and Swedish Paraphrasing Detection Task on Trial Dataset.

Table 24
Comparison of English Prompt 2 Results on Trial Dataset from Gemma, GPT-3.5, Llama3 and GPT-4 for the paraphrasing detection task. Cases highlighted in yellow indicate discrepancies between the model's predictions and the correct labels. Analysis of the table shows consistent challenges for the model in accurately predicting labels, particularly for hallucinations involving pronouns, named entities, natural transformations and conversions. Gemma had the most difficulty across different types of hallucinations. GPT-4 showed the best performance specifically in detecting false pronouns, as the other models failed to recognize this type of hallucination.

| id | type | label | prediction gemma | prediction gpt 3.5 | prediction llama3 | prediction gpt4 |
|---|---|---|---|---|---|---|
| 0 | antonym | hyp1 | hyp1 | hyp1 | hyp1 | hyp1 |
| 1 | negation | hyp1 | hyp1 | hyp1 | hyp1 | hyp1 |
| 2 | antonym | hyp1 | hyp1 | hyp1 | hyp1 | hyp1 |
| 3 | named entity | hyp1 | hyp2 | hyp1 | hyp1 | hyp1 |
| 4 | natural | hyp2 | hyp2 | hyp2 | hyp2 | hyp2 |
| 5 | addition | hyp2 | hyp2 | hyp2 | hyp2 | hyp2 |
| 6 | gender | hyp2 | hyp2 | hyp2 | hyp2 | hyp2 |
| 7 | natural | hyp1 | hyp2 | hyp1 | hyp2 | hyp1 |
| 8 | number | hyp2 | hyp2 | hyp2 | hyp2 | hyp2 |
| 9 | pronoun | hyp1 | hyp2 | hyp2 | hyp2 | hyp1 |
| 10 | pronoun | hyp1 | hyp2 | hyp2 | hyp2 | hyp1 |
| 11 | addition | hyp2 | hyp2 | hyp2 | hyp2 | hyp2 |
| 12 | conversion | hyp1 | hyp2 | hyp1 | hyp1 | hyp1 |
| 13 | natural | hyp2 | hyp2 | hyp1 | hyp2 | hyp2 |
| 14 | named entity | hyp2 | hyp2 | hyp2 | hyp2 | hyp2 |
| 15 | date | hyp1 | hyp1 | hyp1 | hyp1 | hyp1 |

Table 25
Classification Report Gemma of English Prompt 2 Results on Trial Dataset.

|  | Precision | Recall | F1-Score | Support |
|---|---|---|---|---|
| **hyp1** | 1.00 | 0.44 | 0.62 | 9 |
| **hyp2** | 0.58 | 1.00 | 0.74 | 7 |
| **Accuracy** |  |  | 0.69 | 16 |
| **Macro avg** | 0.79 | 0.72 | 0.68 | 16 |
| **Weighted avg** | 0.82 | 0.69 | 0.67 | 16 |

Table 26
Classification Report GPT-3.5 of English Prompt 2 Results on Trial Dataset.

|  | Precision | Recall | F1-Score | Support |
|---|---|---|---|---|
| **hyp1** | 0.88 | 0.78 | 0.82 | 9 |
| **hyp2** | 0.75 | 0.86 | 0.80 | 7 |
| **Accuracy** |  |  | 0.81 | 16 |
| **Macro avg** | 0.81 | 0.82 | 0.81 | 16 |
| **Weighted avg** | 0.82 | 0.81 | 0.81 | 16 |

**Table 27**
Classification Report Llama3 of English Prompt 2 Results on Trial Dataset.

|                  | Precision | Recall | F1-Score | Support |
|------------------|-----------|--------|----------|---------|
| **hyp1**         | 1.00      | 0.67   | 0.80     | 9       |
| **hyp2**         | 0.70      | 1.00   | 0.82     | 7       |
| **Accuracy**     |           |        | 0.81     | 16      |
| **Macro avg**    | 0.85      | 0.83   | 0.81     | 16      |
| **Weighted avg** | 0.87      | 0.81   | 0.81     | 16      |

**Table 28**
Classification Report GPT-4 of English Prompt 2 Results on Trial Dataset.

|                  | Precision | Recall | F1-Score | Support |
|------------------|-----------|--------|----------|---------|
| **hyp1**         | 1.00      | 1.00   | 1.00     | 9       |
| **hyp2**         | 1.00      | 1.00   | 1.00     | 7       |
| **Accuracy**     |           |        | 1.00     | 16      |
| **Macro avg**    | 1.00      | 1.00   | 1.00     | 16      |
| **Weighted avg** | 1.00      | 1.00   | 1.00     | 16      |

**Table 29**
Comparison of Swedish Prompt 1 Results on Trial Dataset from Gemma and GPT-3.5 for the paraphrasing detection task.

| id | type         | label | prediction gemma | prediction gpt 3.5 |
|----|--------------|-------|------------------|--------------------|
| 0  | number       | hyp2  | hyp2             | hyp2               |
| 1  | natural      | hyp1  | hyp2             | hyp2               |
| 2  | named entity | hyp2  | hyp2             | hyp2               |
| 3  | addition     | hyp2  | hyp2             | hyp1               |
| 4  | negation     | hyp1  | hyp1             | hyp1               |
| 5  | gender       | hyp2  | hyp2             | hyp2               |
| 6  | antonym      | hyp2  | hyp2             | hyp2               |
| 7  | negation     | hyp1  | hyp2             | hyp2               |
| 8  | addition     | hyp2  | hyp2             | hyp2               |
| 9  | number       | hyp2  | hyp2             | hyp2               |
| 10 | natural      | hyp1  | hyp2             | hyp1               |
| 11 | addition     | hyp2  | hyp2             | hyp2               |
| 12 | addition     | hyp2  | hyp2             | hyp2               |
| 13 | named entity | hyp1  | hyp2             | hyp2               |
| 14 | named entity | hyp1  | hyp1             | hyp2               |
| 15 | number       | hyp2  | hyp2             | hyp2               |
| 16 | addition     | hyp1  | hyp2             | hyp1               |
| 17 | tense        | hyp2  | hyp2             | hyp2               |
| 18 | pronoun      | hyp1  | hyp2             | hyp2               |
| 19 | date         | hyp1  | hyp1             | hyp1               |

**Table 30**
Classification Report Gemma for Swedish Prompt 1 Results on Trial Dataset.

|              | Precision | Recall | F1-Score | Support |
| ------------ | --------- | ------ | -------- | ------- |
| **hyp1**     | 1.00      | 0.33   | 0.50     | 9       |
| **hyp2**     | 0.65      | 1.00   | 0.79     | 11      |
| **Accuracy** |           |        | 0.70     |         |
| **Macro avg**    | 0.82  | 0.67   | 0.64     | 20      |
| **Weighted avg** | 0.81  | 0.70   | 0.66     | 20      |

**Table 31**
Classification Report GPT-3.5 for Swedish Prompt 1 Results on Trial Dataset.

|              | Precision | Recall | F1-Score | Support |
| ------------ | --------- | ------ | -------- | ------- |
| **hyp1**     | 0.80      | 0.44   | 0.57     | 9       |
| **hyp2**     | 0.67      | 0.91   | 0.77     | 11      |
| **Accuracy** |           |        | 0.70     |         |
| **Macro avg**    | 0.73  | 0.68   | 0.67     | 20      |
| **Weighted avg** | 0.73  | 0.70   | 0.68     | 20      |

**Table 32**
Comparison of Swedish Prompt 2 Results on Trial Dataset from Gemma and GPT-3.5 for the paraphrasing detection task.

| id | type         | label | prediction gemma | prediction gpt 3.5 |
| -- | ------------ | ----- | ---------------- | ------------------ |
| 0  | number       | hyp2  | hyp2             | hyp2               |
| 1  | natural      | hyp1  | hyp2             | hyp2               |
| 2  | named entity | hyp2  | hyp2             | hyp2               |
| 3  | addition     | hyp2  | hyp2             | hyp1               |
| 4  | negation     | hyp1  | hyp1             | hyp1               |
| 5  | gender       | hyp2  | hyp2             | hyp2               |
| 6  | antonym      | hyp2  | hyp2             | hyp2               |
| 7  | negation     | hyp1  | hyp1             | hyp1               |
| 8  | addition     | hyp2  | hyp2             | hyp1               |
| 9  | number       | hyp2  | hyp2             | hyp2               |
| 10 | natural      | hyp1  | hyp2             | hyp1               |
| 11 | addition     | hyp2  | hyp2             | hyp1               |
| 12 | addition     | hyp2  | hyp2             | hyp2               |
| 13 | named entity | hyp1  | hyp2             | hyp2               |
| 14 | named entity | hyp1  | hyp2             | hyp2               |
| 15 | number       | hyp2  | hyp2             | hyp2               |
| 16 | addition     | hyp1  | hyp2             | hyp2               |
| 17 | tense        | hyp2  | hyp2             | hyp2               |
| 18 | pronoun      | hyp1  | hyp2             | hyp2               |
| 19 | date         | hyp1  | hyp2             | hyp2               |

```
1           result = {'label': 'hyp1'}
2
3           src = "The population has declined in some 210 of the 280 municipalities in
        Sweden, mainly in inland central and northern Sweden."
4
5           "In the majority of Sweden's 280 municipalities, the population has gone up."
6           "In the majority of Sweden's 280 municipalities, the population has gone down.
        "
7
8           hyp1
9           hyp2
10
11          if 'declined' in src or 'down' in src:
12              result['label'] = hyp2
13
14          elif 'up' in src:
15              result['label'] = hyp1
16      '''
17
```

**Listing 15:** En_Para_Gen _Gemma_v1

**Figure 12:** Example of generated output from Gemma.

```
1  answer_format = {"label": ""}
2     user_prompt = f'''
3     <start_of_turn>user
4     Given a "src" and two hypotheses "hyp1"
        and "hyp2" your task is to detect
        which of the two hypotheses ("label")
        is not supported by the source.
5     Provide the result in the following
        format: {answer_format}.
6
7     Src: {source}
8     hyp1 : {hyp1}
9     hyp2 : {hyp2}
10
11    <end_of_turn>
12
13    Result:
14    <start_of_turn>model
15    '''
16
```

**Listing 16:** En_Para_Det_Gemma_v1

```
1  answer_format = {"label": ""}
2     user_prompt = f'''
3     <start_of_turn>user
4     Givet en "src" och två hypoteser "hyp1"
        och "hyp2" är din uppgift att upptäcka
        vilken av de två hypoteserna ("label")
        som inte stöds av källan.
5     Ge resultatet i följande format: {
        answer_format}.
6
7     Src: {source}
8     hyp1 : {hyp1}
9     hyp2 : {hyp2}
10
11    <end_of_turn>
12
13    Resultat:
14    <start_of_turn>model
15    '''
16
```

**Listing 17:** Se_Para_Det_Gemma_v1

**Figure 13:** First Prompts used by 'Gemma' for Paraphrasing Detection Task on the test set.

**Table 33**
Classification report for Llama 3 on Trial Dataset.

|              | Precision | Recall | F1-Score | Support |
|--------------|-----------|--------|----------|---------|
| **hyp1**     | 1.00      | 0.50   | 0.67     | 4       |
| **hyp2**     | 0.67      | 1.00   | 0.80     | 4       |
| **Accuracy** |           |        | 0.70     |         |
| **Macro avg**    | 0.83  | 0.75   | 0.73     | 8       |
| **Weighted avg** | 0.83  | 0.75   | 0.73     | 8       |

```
1  answer_format = {"hyp+": "", "hyp-": ""}
2
3  user_prompt = <start_of_turn>user
4    You are a researcher investigating a new phenomenon. You have gathered data
        ({source}) and formulated two competing hypotheses (hyp1: {hyp1}, and hyp2:
        {hyp2}) to explain it.
5    Identify the hypothesis that contradicts the information provided in the given
        source.
6    Provide the result in the following format: {answer_format}.
7    <end_of_turn>
8    Result:
9    <start_of_turn>model
10
```

**Listing 18:** En_Se_Para_Det_Gemma_v2

**Figure 14:** Second Prompts used by 'Gemma' for Paraphrasing Detection Task on the test set.

```
1       answer_format = {"label": ""}
2
3  user_prompt = f'''
4  Given a "src" and two hypotheses "hyp1"
     and "hyp2" your task is to detect
     which of the two hypotheses ("label")
     is not supported by the source.
5  Provide the result in the following
     format: {answer_format}.
6
7  Src: {source}
8  hyp1 : {hyp1}
9  hyp2 : {hyp2}
10
11 Result:
12 '''
13
14
```

**Listing 19:** En_Se_Para_Det_GPT3.5_GPT4_v1

```
1       answer_format = {"label": ""}
2
3  user_prompt = f'''
4  You have gathered data ({source}) and
     formulated two competing hypotheses to
      explain it.
5  hyp1: {hyp1}
6  hyp2: {hyp2})
7  Identify the hypothesis that
     contradicts the information provided
     in the given source.
8
9  Provide the result in the following
     format: {answer_format}.
10
11 Result:
12 '''
13
```

**Listing 20:** En_Se_Para_Det_GPT3.5_GPT4_v2

**Figure 15:** Prompt used by 'GPT-3.5 Turbo' and 'GPT-4' for Paraphrasing Detection Task on the test set.

```
1       Given a "src" and two hypotheses
     "hyp1" and "hyp2" your task is to
     detect which of the two hypotheses ("
     label") is not supported by the source.
2  Provide the result in the following
     format: {answer_format}.
3
4  Src: {source}
5  hyp1 : {hyp1}
6  hyp2 : {hyp2}
7
8  Result:
9    '''
10
```

**Listing 21:** En_Para_Det _Llama3_v1

```
1       Givet en "src" och två hypoteser "
     hyp1" och "hyp2" är din uppgift att
     upptäcka vilken av de två hypoteserna
     ("label") som inte stöds av källan.
2  Ge resultatet i följande format: {
     answer_format}.
3
4  Src: {source}
5  hyp1 : {hyp1}
6  hyp2 : {hyp2}
7
8  Resultat:
9    '''
10
```

**Listing 22:** Se_Para_Det_Llama3_v1

**Figure 16:** First Prompts used by 'Llama 3' for Paraphrasing Detection Task on the test set.

```
1  You are a researcher investigating a new phenomenon.
2  You have gathered data (source) and formulated two competing hypotheses (hyp1 and hyp2
       ) to explain it.
3
4  Identify the hypothesis that contradicts the information provided in the given source.
5  Provide the result in the following format: {answer_format}.
6
7      source: {source}
8      hyp1: {hyp1}
9      hyp2: {hyp2}
10
11     Result:
12
```

Listing 23: En_Se_Para_Det _Llama3_v2

Figure 17: Second Prompt used by 'Llama 3' for Paraphrasing Detection Task on the test set.

**Table 34**

Llama3 exhibits gender bias in the translation detection task.

| Source | hyp1 | hyp2 | Llama3's explanation |
| --- | --- | --- | --- |
| Der Anwalt forderte die Wirtschaftsprüferin auf, Unterlagen vorzulegen. | The lawyer told the male auditor to provide documentation. | The lawyer told the female auditor to provide documentation. | I chose hyp2 as the hallucination because it introduces a new detail that is not present in the source text. The source text only mentions "Wirtschaftsprüferin", which is a gender-neutral term for an auditor. The addition of "female" is an assumption made by the translator, which is not supported by the original text. |
| Der Arbeiter grub ein Loch für die Sekretärin. | The laborer dug a hole for the female secretary. | The laborer dug a hole for the male secretary. | My choice is based on the fact that the original sentence does not mention the gender of the secretary. The source sentence only mentions "die Sekretärin" which is a feminine noun in German, but it does not imply that the laborer dug a hole specifically for a female secretary. The addition of "male" in hyp2 is an unnecessary detail that is not supported by the original sentence. Therefore, I label hyp2 as the hallucination and identify the type of hallucination as addition. |

```
1  user_prompt = f'''
2  Given a "src" and two hypotheses "hyp1" and "hyp2", which are translated into German. Your
       task is to detect which of the two hypotheses is hallucinated ("label") and determine what
        type of hallucination ("type") it is.
3  Please explain your choice.
4  Provide the result in the following format: {"type": "", "label": ""}.
5  Here is an example to illustrate this:
6
7      Src: The days in the summer can lead to problems getting sufficient sleep and associated
       health issues.
8      hyp1: Die Tage im Sommer können zu Problemen führen, genügend Schlaf zu bekommen und damit
        verbundene Gesundheitsprobleme.
9      hyp2: Die sehr langen Tage im Sommer können zu Problemen führen, genügend Schlaf zu
       bekommen und damit verbundene Gesundheitsprobleme.
10     label: hyp2
11     type: addition
12
13 Detect which of the following hypotheses is the hallucination.
14
15     Src: {source}
16     hyp1 : {hyp1}
17     hyp2 : {hyp2}
18
19 Result:
20
```

**Listing 24:** En_De_Trans_Det_llama3_v1, other language pairs contain an example in the associated language

```
1  user_prompt = f'''
2  You are a researcher investigating a new phenomenon. You have gathered data (source) and
       formulated two competing hypotheses (Hyp1 and Hyp2) to explain it. Identify the hypothesis
        that contradicts the information provided in the given source.
3  Provide the result in the following format: {"label": ""}.
4
5  Src: {source}
6  hyp1 : {hyp1}
7  hyp2 : {hyp2}
8
9  Result:
10
```

**Listing 25:** En_De_Trans_Det_llama3_v2, En_Fr_Trans_Det_llama3_v2

**Figure 18:** Prompts Llama3 Translation Detection Task.

**Table 35**
Llama3 struggles with converting measurements in the translation detection task.

| Source | hyp1 | hyp2 | Llama 3's explanation |
| --- | --- | --- | --- |
| Das Zentrum des Erdbebens befand sich etwa 20 km (15 Meilen) nord-nordöstlich von Dillon und etwa 65 km (40 Meilen) südlich von Butte. | The center of the earthquake was about 20 km northeast of Dillon and about 65000 metres south of Butte. | The center of the earthquake was about 20 km northeast of Dillon and about 65000 kilometres south of Butte. | I chose hyp2 as the hallucination because it adds the unit "kilometres" to the distance, which is not present in the original text. The original text uses "km" and "Meilen", which are equivalent units, but hyp2 introduces a new unit "kilometres". This is a type of hallucination where a new information is added to the original text, specifically a unit conversion.' |

**Table 36**
Llama3 struggles with recognizing the different ways dates in the translation detection task.

| Source | hyp1 | hyp2 | Llama3's explanation |
| --- | --- | --- | --- |
| "Perry erklärte, er werde „nach Texas zurückkehren, um das Ergebnis der Vorwahl von heute Abend auszuwerten und zu entscheiden, ob es in diesem Rennen einen weiterführenden Weg für mich gibt"", aber sagte später, er werde im Rennen bleiben und bei den Vorwahlen in South Carolina am 21. Januar antreten." | """Perry stated that he will return to Texas to evaluate the result of tonight's pre-election and decide whether there is a path forward for me in this race,"" but later said he would stay in the race and run for the pre-election in South Carolina on March 21." | """Perry stated that he will return to Texas to evaluate the result of tonight's pre-election and decide whether there is a path forward for me in this race,"" but later said he would stay in the race and run for the pre-election in South Carolina on Jan. 21." | I chose hyp2 as the hallucination because it contains a specific date ("Jan. 21"), which is not present in the original source text. The original text only mentions the date "21. Januar", which is a more general and less specific date. This suggests that hyp2 contains a hallucinated detail. The type of hallucination is time, as it involves the introduction of a specific and precise time element. |

**Table 37**
The prompt immensely influences the outcome of Llama3 for the translation detection task.

| Source | hyp1 | hyp2 | Llama3's explanation |
| --- | --- | --- | --- |
| Der Anwalt forderte die Wirtschaftsprüferin auf, Unterlagen vorzulegen. | The lawyer told the male auditor to provide documentation. | The lawyer told the female auditor to provide documentation. | The source states "Der Anwalt forderte die Wirtschaftsprüferin auf, Unterlagen vorzulegen.", which translates to "The lawyer demanded the female auditor to provide documentation." This implies that the auditor is a woman, contradicting Hyp2 which states that the auditor is a man. result_text 'label': 'hyp2' |
| Der Arbeiter grub ein Loch für die Sekretärin. | The laborer dug a hole for the female secretary. | The laborer dug a hole for the male secretary. | The source states "Der Arbeiter grub ein Loch für die Sekretärin", which means "The laborer dug a hole for the secretary". This clearly indicates that the secretary is a female, as "Sekretärin" is the German word for "female secretary". Therefore, Hyp2, which suggests that the laborer dug a hole for the male secretary, contradicts the information provided in the source. result_text 'label': 'hyp2' |
| Das Zentrum des Erdbebens befand sich etwa 20 km (15 Meilen) nord-nordöstlich von Dillon und etwa 65 km (40 Meilen) südlich von Butte. | The center of the earthquake was about 20 km northeast of Dillon and about 65000 metres south of Butte. | The center of the earthquake was about 20 km northeast of Dillon and about 65000 kilometres south of Butte. | Explanation: The source states that the distance from the earthquake's center to Butte is approximately 40 miles (65 km), not 65,000 kilometres. Hyp2 contradicts this information by stating a much larger distance, while Hyp1 correctly translates the original distance from miles to meters. result_text 'label': 'hyp2' |
| "Perry erklärte, er werde „nach Texas zurückkehren, um das Ergebnis der Vorwahl von heute Abend auszuwerten und zu entscheiden, ob es in diesem Rennen einen weiterführenden Weg für mich gibt"", aber sagte später, er werde im Rennen bleiben und bei den Vorwahlen in South Carolina am 21. Januar antreten." | """Perry stated that he will return to Texas to evaluate the result of tonight's pre-election and decide whether there is a path forward for me in this race,"" but later said he would stay in the race and run for the pre-election in South Carolina on March 21." | """Perry stated that he will return to Texas to evaluate the result of tonight's pre-election and decide whether there is a path forward for me in this race,"" but later said he would stay in the race and run for the pre-election in South Carolina on Jan. 21." | Explanation: The source states that the event in question is on "heute Abend" (tonight), which corresponds to January 21. However, Hyp1 states that the event is on "March 21", which contradicts the information provided in the source. result_text 'label': 'hyp1' |

Table 38
The count of failed examples for GPT-3.5 Turb and GPT-4 in the translation detection task.

| Language | GPT-4 | | GPT-3.5 | |
| --- | --- | --- | --- | --- |
| | prompt1 | prompt2 | prompt1 | prompt2 |
| de-en | 2 | 1 | 0 | 0 |
| en-de | 8 | 15 | 1 | 1 |
| en-fr | 2 | 7 | 0 | 1 |
| fr-en | 7 | 3 | 1 | 2 |

```
user_prompt = f'''
Given a "src" and two hypotheses "hyp1" and
    "hyp2",
your task is to first translate "hyp1" and
    "hyp2"
back to the language of "src". Then,
    compare these
translations with "src" to detect which of
    the two
hypotheses is hallucinated ("label") and
    what type of
hallucination ("type") it is. Provide the
    result in the
following format: {answer_format}.

Src: {source}
hyp1 : {hyp1}
hyp2 : {hyp2}
Result:
'''
```

**Listing 26:** En_De_Trans_Det_gemma_v1, En_Fr_Trans_Det_gemma_v1

```
user_prompt = f'''
<start_of_turn>user
You are a researcher investigating a new
    phenomenon. You have gathered data (
    source) and formulated two competing
    hypotheses (hyp1 and hyp2) to explain
    it. Identify the hypothesis that
    contradicts the information provided
    in the given source. Provide the
    result in the following format: {
    answer_format}. ONLY ANSWER IN THAT
    FORMAT!

source: {source}
hyp1 : {hyp1}
hyp2 : {hyp2}

<end_of_turn>

Result:
<start_of_turn>model

```

**Listing 27:** De_En_Trans_Det_gemma_v2, En_De_Trans_Det_gemma_v2, En_Fr_Trans_Det_gemma_v2, Fr_En_Trans_Det_gemma_v2

**Figure 19:** Prompts Gemma Translation Detection Task.

```
        Given a "src" and two hypotheses "hyp1" and "hyp2" your task is to detect which
of the two hypotheses ("label") is not supported by the source.
        Provide the result in the following format:
        {answer_format}.

        Src: {source}

        hyp1: {hyp1}

        hyp2: {hyp2}

```

**Listing 28:** majority_vote_cross_model_result_en

**Figure 20:** The prompts used for paraphrasing task in cross-model.

**Table 39**
Samples for which GPT-4 failed to assign labels in the translation detection task.

| Lang | Source | hyp1 | hyp2 | Our explanation |
|---|---|---|---|---|
| de-en | Die Mittel könnte man für hochwassersichere Häuser, eine bessere Wasserverwaltung und Nutzpflanzendiversifizierung verwenden. | The funds could be used for more waterproof houses, better water management and crop diversification. | The funds could be used for more flood-proof houses, better water management and crop diversification. | Different interpretations of flood-proof |
| de-en | Es zeigt 362 verschiedene alte Holzarten, Büsche und 236 verschiedene Obstbaumarten. | It shows 362 different old species of wood, bushes and 236 different species of fruit trees. | It showed 362 different old species of wood, bushes and 236 different species of fruit trees. | wrong tense |
| en-de | The world has over 5,000 different languages, more than twenty with 50 million or more speakers. | Die Welt hat mehr als 5.000 verschiedene Sprachen, darunter mehr als zwanzig mit 50 Millionen oder mehr Sprechern. | Die Welt hat mehr als 5.000 verschiedene Sprachen, mehr als zwanzig mit 50 Millionen oder mehr Sprechern. | missing filler word |
| en-de | 1i Productions is an American board game publisher. It was founded in 2004 by Colin Byrne, William and Jenna. | Es wurde 2004 von Colin Byrne, William und Jenna gegründet und ist ein amerikanischer Brettspielverlag. | 1i Productions ist ein amerikanischer Brettspieleverlag. Er wurde von Colin Byrne, Wiliam and Jenna im Jahr 2004 gegründet. | Name of the publisher missing |
| en-de | Mats Wilander defeats Anders Järryd, 6 – 4, 3 – 6, 7 - 5. | Mat Wilander besiegt Anders Järryd, 6:4, 3:6, 7:5. | Mats Wilander schlägt Anders Järryd 6:4, 3:6, 7:5. | different (but still correct) translations of defeat |
| en-de | They have feet with scales and claws, they lay eggs, and they walk on their two back legs like a T-Rex. | Sie haben Füße mit Schalen und Nägeln, sie legen Eier und sie gehen auf ihren beiden Rückenbeinen wie ein T-Rex. | Sie haben Füße mit Schalen und Nägeln, sie legen Eier und sie gehen auf ihren beiden Hinterbeinen wie ein T-Rex. | Hind legs was translated incorrectly from back (body part) |
| en-de | The NSA has its own internal data format that tracks both ends of a communication, and if it says, this communication came from America, they can tell Congress how many of those communications they have today, right now. | Die NSA hat ein eigenes internes Datenformat, das beide Enden einer Kommunikation verfolgt, und wenn sie sagt, diese Mitteilung kam aus Amerika, können sie dem Kongress sagen, wie viele dieser Mitteilungen es heute haben, gerade jetzt. | Die NSA hat ein eigenes internes Datenformat, das beide Enden einer Kommunikation verfolgt, und wenn sie sagt, diese Mitteilung kam aus Amerika, können sie dem Kongress sagen, wie viele dieser Mitteilungen sie heute haben, gerade jetzt. | Wrong pronoun "it" instead of "they" |
| en-de | In 2014 the site launched iOS and Android applications for product search; product features include interactive video product reviews with live question-and-answer sessions. | 20124 brachten iOS und Android Applikationen zur Produktsuche heraus; Produktfeatures beinhalten interaktive Video-Produktreviews mit live Frage- und Antwort-Sessions. | Im Jahr 2014 startete die Website iOS und Android - Anwendungen für die Produktsuche. Zu den Produktfunktionen gehören interaktive Videoproduktbewertungen mit live Fragen und Antworten. | The year is messed up |